\newcommand{\nuc}{\newcommand}
\nuc{\Sg}{\Sigma}

\nuc{\EBN}{\begin{equation}}
\nuc{\EEN}{\end{equation}}
\newcommand{\EB}{\begin{eqnarray*}}
\newcommand{\EE}{\end{eqnarray*}}

\nuc{\mbo}{$\mbox{ }$}
\nuc{\mbh}{\mbo\mbo}
\nuc{\lam}{\lambda}

\nuc{\kmle}{k-MLE }
\nuc{\kmlE}{k-MLE}
\nuc{\Omx}{\Omega^{\times}}

\nuc{\om}{\omega}
\nuc{\Xbf}{{\bf X}}
\nuc{\Xbfnk}{\Xbf_{nk}}
\nuc{\Ybfn}{{\bf Y}_n}
\nuc{\Ebfnk}{{\bf E}_{nk}}

\nuc{\Qnk}{Q_{nk}}
\nuc{\Rnk}{R_{nk}}
\nuc{\Fbfn}{{\bf F}_n}
\nuc{\Gbfn}{{\bf G}_n}

\nuc{\T}{^\top}
\nuc{\summb}[2]{\mbox{$\sum_{#1}^{#2}$}}
\nuc{\pk}{p_k}
\nuc{\Ybfqnk}{{\bf Y}_{Qnk}}

\nuc{\udl}[1]{\underline{#1}}
\nuc{\nn}{\nonumber}

\nuc{\ep}{\epsilon}

\nuc{\wox}[1]{\frac{1}{#1}}

\nuc{\enk}{e_{nk}}
\nuc{\dadb}[2]{\frac{\pde #1}{\pde #2}}
\nuc{\calk}{{\cal L}_k}
\nuc{\enkt}{e_{nkt}}
\nuc{\half}{\frac{1}{2}}

\nuc{\EQ}{\begin{displaymath}}
\nuc{\EN}{\end{displaymath}}
\newcommand{\EQN}{\begin{equation}}
\newcommand{\ENN}{\end{equation}}

\nuc{\Om}{\Omega}
\nuc{\Omk}{\Omega_k}
\nuc{\Omhk}{\hat{\Omega}_k}
\nuc{\pde}{\partial }

\nuc{\upm}{^{-1}}
\nuc{\Ykmr}{Y_{k-r}}
\nuc{\Yupntr}{Y^{(n)}_{t-r}}
\nuc{\Thtr}{\Theta_r}
\nuc{\Thtkr}{\Theta_{k,r}}
\nuc{\sumpT}{\summb{t=p+1}{T}}
\nuc{\Snk}{S_{n,k}}
\nuc{\ul}{\underline}

\documentclass[letterpaper,10pt,journal,final]{IEEEtran}
\usepackage[utf8]{inputenc}
\usepackage[T1]{fontenc}
\usepackage[scaled=0.95]{inconsolata}

\usepackage{latexsym,amsmath,amssymb,amsfonts,mathrsfs,
            mathtools, amsthm}
\usepackage{arydshln}   
\usepackage{empheq}     
\usepackage{enumitem}   
\usepackage{booktabs,makecell}   

\usepackage{algorithm, algpseudocode}
\makeatletter
\renewcommand{\ALG@beginalgorithmic}{\small}
\makeatother

\usepackage{float, graphicx, caption, subcaption}
\usepackage[usenames,dvipsnames]{xcolor}
\usepackage[colorlinks=true,linkcolor=magenta,citecolor=blue,
            urlcolor=cyan,filecolor=red]{hyperref}
\usepackage{cite}

\usepackage{comment}

\newtheorem{theorem}{Theorem}
\newtheorem{lemma}[theorem]{Lemma}
\newtheorem{proposition}[theorem]{Proposition}

\theoremstyle{definition}
\newtheorem{definition}{Definition}
\newtheorem{assumption}{Assumption}
\theoremstyle{remark}

\newcommand{\argmin}{\operatorname{arg\,min}}

\title{k-MLE, k-Bregman, k-VARs: \\ Theory, Convergence, Computation}

\author{Zuogong~Yue$^{1,2}$,~\IEEEmembership{Member, IEEE} and
  Victor~Solo$^{2*}$,~\IEEEmembership{Life Fellow, IEEE}%
  \thanks{$^1$Z. Yue is with the
    School of Artificial Intelligence and Automation,
    Huazhong University of Science and Technology,
    Wuhan, Hubei 430074 CHINA}%
  \thanks{$^2$Z. Yue was, and V. Solo is, with
    School of Electrical Engineering and Telecommunications,
    University of New South Wales,
    Kensington, NSW 2052 AUSTRALIA}%
  \thanks{$^*$For correspondence, \texttt{v.solo@unsw.edu.au}}
}

\begin{document}

\maketitle
\thispagestyle{empty}
\pagestyle{empty}

\begin{abstract}
  In this paper we develop a general approach to hard clustering which we
  call k-MLE and provide a general convergence result for it. 
%
  Unlike other hard clustering generalizations of k-means, which are
  based on distance or divergence, k-MLE is based on likelihood and thus
  has a far greater range of application. 
  We show that `k-Bregman'
  clustering is a special case of k-MLE and thus provide, for the first
  time, a complete proof of convergence for k-Bregman clustering. 
  We give a further application: k-VARs for clustering
 vector autocorrelated/autoregressive time series.
%
It does not admit a Bregman divergence interpretation.
We provide simulations and real data application as well as convergence results.
%
\end{abstract}


\section{Introduction}
\label{sec:introduction}

\IEEEPARstart{C}{lustering}  is a method for partitioning
a collection of data vectors into groups (called clusters).
It is a common technique for  data mining
and exploratory statistical  analysis. 
It has been widely used in many fields, for
instance, pattern recognition\cite{Duda2001}, image
processing\cite{Sonka2015}, computer vision\cite{Szeliski2011},
bioinformatics\cite{Jones2004}, and etc. The study of cluster analysis has
a long history, going back to  anthropology in the 1930s
\cite{Driver1932}.

k-MLE.  
There are a wide range of clustering methods
\cite{Duda2001},\cite{Hastie2009},\cite{Everitt2011,Hennig2016}
with new ones still emerging
e.g. \cite{PELK05},\cite{CHEN15}.
There are two types: hard clustering methods and soft clustering methods.
In hard clustering the groups are completely separate.
In soft clustering the groups may overlap somewhat.
So far hard clustering algorithms are all based
 on similarity measures,
typically a distance or pseudo-distance \cite{Banerjee2005}.
Soft clustering methods are constructed in three ways:
(i) based on likelihood via mixture models \cite{Duda2001} fitted with the EM algorithm; 
(ii) based on fuzzy methods \cite{MIH17};
(iii) based on regularization e.g., with the entropy penalty \cite{KARA94}.
In this paper we develop a new and general approach to hard clustering
but based on the `classification' likelihood \cite{Banfield1993} not similarity measures.
We prove a general convergence result. 
And we show that a class of existing similarity/divergence 
methods \cite{Banerjee2005} form a special case
which we rename  as k-Bregman.
As an application we develop
a new method (k-VARs) for clustering autocorrelated time series.


Convergence. 
For the k-means algorithm with general divergence measures,
 the definitive work on convergence is the seminal paper of
\cite{Selim1984}. The much later independent
work of \cite{Bottou1995}, while useful, is much less general.
We will also discuss the work of \cite{Banerjee2005} which
also includes convergence analyses for k-Bregman, which we show are incomplete.

Model Selection. 
The problem of choosing the number of clusters is
a challenging problem which garners ongoing interest.
Important methods include 
BIC \cite{Schwarz1978} and
the gap method \cite{YANY07}.
Since k-MLE is based on likelihood, it is straightforward
to develop a BIC criterion for joint choice of
model order and number of clusters. This is illustrated with k-VARs.

Time Series. 
In many applications
the data to be clustered
are autocorrelated time series, e.g., stock
prices, locations of mobile robots, speech signals, ECG, etc. Statistically
efficient clustering of such temporally autocorrelated data demands an
integration of time-series structure into the clustering methodology. 
There
is already a significant literature on extending clustering algorithms for
time series, e.g.,
k-means integrated with \emph{DTW
      barycenter averaging} (k-DBA) \cite{Petitjean2011}, clustering via
    ARMA mixtures\cite{Xiong2004}, \emph{k-Spectral Centroid} (k-SC)
    algorithm \cite{Yang2011}, \emph{k-Shape}
    \cite{Paparrizos2015,Paparrizos2017}, \emph{shapelet} based methods
    \cite{Ulanova2015,SiyouFotso2020}, deep learning based methods
    \cite{Madiraju2018}, etc. Some methods are only available
    for univariate time series.
%
However some of these `time series' clustering algorithms
ignore the autocorrelation feature and so perform poorly in practice,
when it is usually present.

The remainder of the paper is organized as follows.
The derivation of k-MLE is presented in section II.
In section III a convergence analysis is given.
In section IV, k-MLE is applied to 
clustering autoregressive time series
yielding the k-VARs algorithm.
The section also covers computation,
convergence and model selection.
Application to data and simulations are in section V.
Section VI contains conclusions. There is one appendix containing some proofs.

%
%

%
%
%

\section{k-MLE: A General Tool for Clustering}
\label{sec:kmle-theory}

\subsection{Background}
\label{subsec:k-mle-b}

Given independently distributed (i.d.) data vectors (of dimension $d$\,)
$x_n,n=1,\dots,N$, each belonging to one of $K$ clusters,
$C_1,\dots,C_K$, we introduce the label matrix or binary cluster
membership array $\tau=[\tau_{n,k}]_{N \times K}$
\begin{equation}
  \label{eq:tau}
  \tau_{n,k} =
  \begin{cases}
    1 & \text{if } x_n \in \text{cluster } k, \\
    0 & \text{otherwise}.
  \end{cases}
\end{equation}
The clustering problem is to estimate $\tau_{N\times K}$ from the
$d\times N$ data matrix $X_1^N =[x_1,\cdots,x_N]$. Note that there are
only finitely many $\tau$'s. We denote the $k^{th}$ column of
$\tau$ by $\tau_k$.

Clustering algorithms are often categorised in various ways:
partition methods versus
hierarchical methods;
hard clustering, where each data vector
is assigned to only one cluster, versus soft (a.k.a. fuzzy) clustering, where
each data vector may belong to more than one cluster;
similarity based versus model based.
Here we introduce a new division: between deterministic label clustering
(DL-C) and stochastic label clustering (SL-C). DL-C treats the
membership variables as deterministic, whereas SL-C treats them as
random. We note that SL-C always produces soft clustering whereas DL-C
can produce either hard clustering or soft clustering.

The classic DL-C method is \emph{k-means}\cite{Duda2001}, where clustering is
achieved by minimizing the criterion
\begin{displaymath}
  J = J(\tau, \mu) = \textstyle\sum_{n=1}^N \sum_{k=1}^K \tau_{n,k} d(x_n, \mu_k).
\end{displaymath}
w.r.t. $\tau,\mu$. Here $\mu_k$ is a cluster centre and $d(x,\mu)$ is a
distance or similarity measure, i.e. it has two properties: (i)
positivity $d(x,\mu)\geq 0$; (ii) uniqueness $d(x,\mu)=0$ iff $x=\mu$.
Sometimes symmetry is useful: (iii) $d(x,\mu)=d(\mu,x)$. k-means uses
Euclidean distance but other distance measures are also used, e.g.
\emph{k-medoids} \cite{Duda2001}.

The classic SL-C method is based on a mixture model \cite{Duda2001} in
which the $\tau_{nk}$ are independently distributed with
$P(\tau_{nk}=1)=\pi_k$.

Historically almost all hard clustering methods have been based on similarity
measures whereas almost all soft clustering methods are based on models.
This seems to have fostered the impression that hard clustering methods
could not be model based. But we will see that k-MLE is a DL-C method
but is model based. This suggests that a deeper division is between DL
and SL rather than between similarity based and model based methods.
This paper concentrates on DL methods.

\subsection{k-MLE Formulation} 
\label{subsec:k-mle-d}

Suppose that $x_n$ are i.d. with those in cluster $k$ having density
$p(x_n;\theta_k)$ where $\theta_k \in \Omega$ is $q$-dimensional. Then the
joint density or likelihood of the data in cluster $k$ is
$\prod_{n=1}^N \left[ p(x_n; \theta_k) \right]^{\tau_{nk}}$
with corresponding log-likelihood
\begin{equation}
  \label{eq:lld-single}
  \textstyle
  \lam(\tau_k, \theta_k) = \sum_{n=1}^N \tau_{n,k} \ell(x_n, \theta_k)
  = \sum_{n \in C_k} \ell(x_n, \theta_k),
\end{equation}
where $\ell(x_n, \theta_k) = \mathrm{ln}\, p(x_n; \theta_k)$. Denoting
$\Theta = [\theta_1, \dots, \theta_K]$, the joint log-likelihood
of all the data is then
\begin{equation}
  \label{eq:lld}
  \textstyle
    \mathcal{L}(\tau, \Theta)
    = \sum_{k=1}^K \lam(\tau_k, \theta_k)
    = \sum_{n=1}^N \sum_{k=1}^K \tau_{n,k} \ell(x_n, \theta_k).
\end{equation}
We call this the deterministic label clustering likelihood (DL-CL). The
corresponding clustering likelihood for a SL approach would be called a
stochastic label clustering likelihood (SL-CL).

We now note two important properties of the DL-CL which now follow
immediately.
\begin{itemize}
\item Property P1: $\mathcal{L}(\tau,\Theta)$ is linear in $\tau$.
\item Property P2: $\mathcal{L}(\tau,\Theta)$ is separable in
  $(\tau,\Theta)$.
\end{itemize}
%
Separability means that e.g. maximizing over $\Theta$ can be done
by separately maximizing each cluster-specific likelihood.

The DL-CL depends on binary and analog (i.e. continuous valued)
parameters. This leads to a hybrid maximum likelihood estimation (MLE)
problem as follows.
\begin{definition}[k-MLE problem]
  \label{def:k-mle-problem}
  \begin{equation}
    \label{eq:k-mle}
    \begin{array}{r@{\;}l}
      &\underset{\Theta\in \Omega^{\times}, \tau \in
        \mathcal{B}}{\mathrm{maximize}}\;
      \mathcal{L}(\tau, \Theta)\\
      &\mathcal{B} =
        \big\{
        \begin{array}[t]{@{}r@{\;}l}
          \tau : & \sum_{k=1}^K \tau_{n,k} = 1, \tau_{n,k} = 0 \text{ or } 1,\\
                 & 1 \leq k \leq K, 1 \leq n \leq N \big\},
        \end{array}
    \end{array}
  \end{equation}
  where $\Theta \in \Omega^{\times}$ means $\theta_k \in \Omega$, $1
  \leq k \leq K$.
\end{definition}

For this optimization to be well defined we need to introduce some assumptions.
\begin{assumption}
  \label{assump:compact-Om-cont-ell}
  ~
  \begin{enumerate}[label=(\alph*)]
  \item $\Omega$ is an open subset of $\mathbb{R}^q$ with boundary $\partial\Omega$.
  The closure $\Omega\cup\partial\Omega$ is bounded (which thus means the closure is compact).
  \item $\ell(x_n, \theta)$ is continuous for $\theta \in \Omega$.
  \end{enumerate}
\end{assumption}
Note that (a) and (b) ensure that $\ell(x_n, \theta)$ is bounded.
Since $\mathcal{B}$ is a discrete, bounded set, then under Assumption~\ref{assump:compact-Om-cont-ell}, the k-MLE
problem has at least one solution.

The k-MLE algorithm solves the hybrid k-MLE problem by cyclic
ascent\footnote{Cyclic optimization is such a simple idea that it is
  repeatedly reinvented and renamed as if it was new.} (a.k.a.
coordinate ascent\cite{Luenberger2008}).
\begin{definition}[k-MLE algorithm]
  \label{def:k-mle-algor}
  Denote the iteration index by $m$. We assume that initial parameter
  estimates $\theta_k^{(0)}$ are available for each cluster.
  \begin{itemize}
    \setlength\itemsep{.5em}
  \item \underline{$\tau$-step}: given $\tau^{(m)}$ and $\Theta=\Theta^{(m)}$
    get $\tau^{(m+1)}$. Since $\Theta$ is given, we need only maximize,
    for each $n$, $\sum_{k=1}^K \tau_{n,k} \ell(x_n, \theta_k)$ w.r.t.
    $\tau_n=[\tau_{n,1},\dots,\tau_{n,K}]^T$. Denote by $k_n^*$, the index
    $k$ for which the log-likelihood $\ell(x_n,\theta_k)$ is the largest.
    Then
    \begin{equation}
      \label{eq:tau-step}
      \tau_{n,k} =
      \begin{cases}
        1 & \text{if } k = k_n^*, \\
        0 & \text{otherwise},
      \end{cases}
    \end{equation}
    and we set $\tau^{m+1} = [\tau_{n,k}]_{N \times K}$. If
    $\mathcal{L}(\tau^{(m+1)}, \Theta^{(m)}) = \mathcal{L}(\tau^{(m)},
    \Theta^{(m)})$, stop.

  \item \underline{$\Theta$-step}: given $\Theta^{(m)}, \tau =
    \tau^{(m+1)}$ get $\Theta^{(m+1)}$.
    \begin{equation}
      \label{eq:theta-step}
      \theta_k^{(m+1)} = \underset{\theta \in
        \Omega}{\operatorname{arg\,max}}\: \lambda(\tau_k, \theta),
      \quad 1 \leq k \leq K.
    \end{equation}
    This just entails solving an MLE problem for each cluster, using the
    data in that cluster. If
    $\mathcal{L}(\tau^{(m+1)}, \Theta^{(m+1)}) = \mathcal{L}(\tau^{(m+1)},
    \Theta^{(m)})$, stop.
  \end{itemize}
\end{definition}
Notice there is no requirement that the log-density is a similarity
measure. So this provides a considerable generalization of DL algorithms
such as k-means.

After the second author had developed this method we discovered it is
not new. In \cite{Banfield1993} the DL-CL likelihood appears under the
name `classification likelihood'. But it is only presented in the
context of a multivariate Gaussian density. When the densities are
multivariate spherical Gaussian then k-MLE reduces to k-means. No
algorithm details are given in \cite{Banfield1993}.

The DL-CL has also been developed (in a different direction) by
\cite{Zhong2003}  under the name
`model-based k-means'.
It is derived heuristically from a SL-CL and an
algorithm is given without recognizing that it is cyclic ascent.
There is no convergence analysis and no tuning parameter selection.
\cite{Zhong2003} attributes the method to \cite{KRNS97},\cite{BISW02}. The
first reference only deals with two clusters and has no convergence analysis 
or tuning parameter selection.
The second reference fits a mixture model with an EM algorithm and so is not a DL-C method.




\subsection{k-Bregman}
\label{subsec:k-breg}

In a seminal piece of work \cite{Banerjee2005} showed `there exists a
unique Bregman divergence corresponding to every regular exponential
family'. Then, using a notion of Bregman information, they developed a
hard clustering algorithm based on Bregman divergence.
It is not noted in \cite{Banerjee2005} but their algorithm is a
coordinate descent algorithm \cite{Luenberger2008}.
Further \cite{Banerjee2005} provided a convergence result, which we will
see below leaves major questions unanswered.

We now show that the Bregman hard clustering algorithm for exponential
families is a special case of k-MLE; we henceforth refer to it as
\emph{k-Bregman}. It is shown in \cite{Banerjee2005} that if $p(x;\theta)$ is a
\emph{regular exponential family}\footnote{For expontial families in
  general, the same result is remarked in \cite{Forster2002}.} it has a
unique decomposition
\begin{equation}
  \label{eq:bregman-prob}
  \ln p(x; \theta)=- d_{\phi} (x; \mu(\theta)) + \ln (b_{\phi}(x))
\end{equation}
where $d_{\phi}(x;\mu(\theta)) \geq 0$ is the Bregman divergence, $\phi$ is
the conjugate function of $\psi$, $\mu(\theta) = \nabla \psi(\theta)$ is
the expectation parameter corresponding to $\theta$, and the reminder
term $b_{\phi}(x)$ does not depend on the parameter $\theta$ and is a
normalization factor for the probability density. Thus the
log-likelihood for the whole data can be written as
\begin{align}
  \label{eq:bregman-lld}
  \mathcal{L}(\tau, \Theta)
  &= \textstyle - \sum_{n=1}^N \sum_{k=1}^K \tau_{n,k} d_{\phi}(x_n; \theta_k)
    + \zeta(X_1^n),\\
  \label{eq:bregman-reminders}
  \zeta(X_1^n)
  &= \textstyle\sum_{n=1}^N \sum_{k=1}^K \tau_{n,k} \ln(b_{\phi}(x_n)).
\end{align}
However, by the property of $\tau_{n,k}$, the remainder term
$\zeta(X_1^n)$ reduces to $\sum_{n=1}^N \ln(b_{\phi}(x_n))$ and can be
dropped. Therefore for an exponential family k-MLE reduces to the
k-Bregman similarity based method.

\subsection{Applications} 
\label{subsec:k-vars}

The possibilities for k-MLE are extensive.
Below we develop
a new algorithm for clustering vector time series which is not
similarity based; we also prove its convergence. 
We previously developed a scalar version \cite{SOLV20c} by taking a limit in an AR mixture
model. That limiting argument can be extended to the vector case but
yields a different algorithm which will be discussed elsewhere.
%

More generally, k-MLE can handle any data type e.g. multi-modal data,
hybrid data with both continuous valued and discrete valued 
measurements; spatio-temporal data; temporal data sampled on mixed time-scales;
multi-subject data and so on.
\section{k-MLE Convergence}
\label{sec:k-mle-c}

This section develops a convergence analysis of the k-MLE algorithm. Our
approach builds on the seminal work of \cite{Selim1984}. Although
\cite{Selim1984} uses dissimilarity where we use log-likelihood, large
parts of their development make no use of the dissimilarity properties.
We also note the work of \cite{Bottou1995} (independent of
\cite{Selim1984}) which, while providing valuable additional insight on
k-means, is much less general in scope and results than
\cite{Selim1984}.

Note that \cite{Selim1984} deals with minima whereas we deal with maxima.
This is a trivial difference. However partly because of our likelihood
framework, we need to rework some of the internal proof developments in
\cite{Selim1984}. Because there is only partial novelty here, we put
most of those proofs in the appendix. However we also need non-trivial
new results involving new conditions.

Let us note immediately from the construction of the k-MLE algorithm
that
\begin{displaymath}
  \begin{array}{r@{\;}l}
    \mathcal{L}^{(m+1)} &= \mathcal{L}(\tau^{(m+1)}, \Theta^{(m+1)})
                          \geq \mathcal{L}(\tau^{(m+1)}, \Theta^{(m)}) \\
                        &\geq \mathcal{L}(\tau^{(m)}, \Theta^{(m)})
                          = \mathcal{L}^{(m)}.
  \end{array}
\end{displaymath}
This ascent property, i.e. non-decrease of the log-likelihood, is
necessary but far from sufficient for convergence of cyclic ascent. If
the log-likelihood is bounded then the ascent property ensures, via the
monotonicity, that the sequence $\mathcal{L}^{(m)}$ has a limit point to which
it converges. But this 
 \udl{criterion} \udl{convergence} 
does not imply 
 \udl{parameter} \udl{convergence}
 i.e.
that $\tau^{(m)}, \Theta^{(m)}$ have
limit points or converge to any of them if they exist. Unfortunately
both \cite{Zhong2003} and \cite{Banerjee2005} mistakenly assert that
criterion convergence implies parameter convergence. In fact
the major/hard part of a proof of convergence is getting from criterion convergence
 to parameter convergence as we now show.

The parameter convergence analysis is in three parts. In the first part we show
convergence in a finite number of steps to what we (c.f.
\cite{Selim1984}) call a partial maximum. In the second part further
study is made to find conditions under which a partial maximum is a
local maximum. In the third part this equivalence is established under
MLE uniqueness.

\subsection{Convergence to Partial Maxima}
\label{subsec:prog-p-partial-opt}

It turns out to be convenient to consider a `relaxed' or purely analog
version of the k-MLE problem in which the membership variables are
analog lying in $[0,1]$. This is formalised by introducing an
\emph{analog constraint set}.

\begin{definition}[analog constraint set]
  \label{def:analog-constr-set}
  \begin{equation}
    \label{eq:analog-constr-set}
    \begin{array}{r@{\;}l}
      \mathcal{A} =
      \big\{
      \tau : & \sum_{k=1}^K \tau_{n,k} = 1, \;
               0 \leq \tau_{n,k} \leq 1, \\
             & 1 \leq k \leq K, 1 \leq n \leq N
      \big\}.
    \end{array}
  \end{equation}
\end{definition}
Note that $\mathcal{A}$ is bounded and $\mathcal{L}(\tau, \Theta)$ is
well defined and bounded for $\tau \in \mathcal{A}, \Theta \in \Omx$.
We can thus introduce the:
\begin{definition}[concentrated\footnote{This is standard statistical
    and econometric terminology.} log-likelihood]
  \label{def:concent-lld}
  \begin{equation}
    \label{eq:concent-lld}
    F(\tau) = \underset{\Theta \in \Omx}{\mathrm{max}} \;
    \mathcal{L}(\tau, \Theta) \quad \text{for } \tau \in \mathcal{A}.
  \end{equation}
\end{definition}

\begin{lemma}[\cite{Selim1984}]
  \label{lem:convex-Ftau}
  $F(\tau)$ is convex.
\end{lemma}
\begin{proof}
  See Appendix~\ref{appdix:proof-lem-convex-Ftau}.
\end{proof}

\begin{lemma}[Theorem~2\cite{Selim1984}]
  \label{lem:extreme-pts}
  The extreme points of $\mathcal{A}$ satisfy the binary constraints
  $\mathcal{B}$ of the k-MLE problem.
\end{lemma}
\begin{proof}
  Elementary.
\end{proof}

We can now introduce the concentrated k-MLE problem:
\begin{definition}[k-MLE-c problem]
  \label{def:k-mle-c-problem}
  \begin{equation}
    \label{eq:k-mle-c}
    \underset{\tau \in \mathcal{A}}{\mathrm{max}} \; F(\tau).
  \end{equation}
\end{definition}

\begin{proposition}[\cite{Selim1984}]
  \label{thm:prob-same-sol}
  The k-MLE-c problem and the k-MLE problem have the same solution set.
\end{proposition}
\begin{proof}
  See Appendix~\ref{appdix:proof-thm-prob-same-sol}.
\end{proof}

\begin{definition}[Partial Maximum \cite{Selim1984}]
  \label{def:partial-maximum}
  A point $(\tau^*, \Theta^*)$ is called a \emph{partial maximum} for
  the k-MLE problem if it satisfies
  \begin{displaymath}
    \begin{array}{r@{\;}l}
      \mathcal{L}(\tau^*, \Theta^*) &\geq \mathcal{L}(\tau, \Theta^*)
      \quad \text{for all } \tau \in \mathcal{A}, \text{ and } \\
      \mathcal{L}(\tau^*, \Theta^*) &\geq \mathcal{L}(\tau^*, \Theta)
      \quad \text{for all } \Theta \in \Omx.
    \end{array}
  \end{displaymath}
\end{definition}

\begin{theorem} 
  \label{thm:conv-parti-max-finite}
  The iterates of the k-MLE algorithm converge to a partial maximum for
  the k-MLE problem in a finite number of steps.
\end{theorem}
\begin{proof}
  See Appendix~\ref{appdix:proof-thm-conv-parti-max-finite}.
\end{proof}

While Theorem~\ref{thm:conv-parti-max-finite} is encouraging, the
problem is that, a partial maximum may not be a local maximum;
\cite{Selim1984} gives an example. We discuss this further in detail in
the next section where we develop conditions for local maxima.

The papers \cite{Banerjee2005} and \cite{Zhong2003} claim to show
convergence to a local optimum, but what they show is that the algorithm
stops in a finite number of steps. They do not obtain
Theorem~\ref{thm:conv-parti-max-finite} because they do not have the
notion of partial maximum which is a prerequisite for a local maximum.

\subsection{Convergence to Local Maxima} 
\label{subsec:conv-part-optima}


We introduce the
\begin{definition}[maximizing solution set]
  \label{def:max-sol-set}
  \begin{equation}
    \label{eq:max-sol-set}
    M(\tau) =
    \big\{
    \Theta_\tau \in \Omx:
    \Theta_{\tau} = \underset{\Theta \in \Omega}{\mathrm{arg\,max}}\;
    \mathcal{L}(\tau, \Theta), \tau \in \mathcal{B}
    \big\}.
  \end{equation}
\end{definition}
Since there are finitely many $\tau$'s, there are finitely many
$M(\tau)$'s. Next, recalling Property~P2, for fixed $\tau$,
$\mathcal{L}(\tau,\Theta)$ is separable. Thus
\begin{equation}
  \label{eq:M-struct}
  \begin{array}{r@{\;}l}
    M(\tau) &= \big\{
              \Theta_{\tau} = [\theta_{1,\tau_1}, \dots, \theta_{K,\tau_K}]
              \big\} \\
            &= M_1(\tau_1) \times \cdots \times M_K(\tau_K) \\
    M_k(\tau_k) &= \big\{
                  \theta_{k,\tau_k} \in \Omega:
                  \theta_{k,\tau_k} = \underset{\theta \in
                  \Omega}{\mathrm{arg\,max}}\; \lambda(\tau_k,\theta)
                  \big\},
  \end{array}
\end{equation}
where the operator ``$\times$'' is the Cartesian product.
We say that $M(\tau)$ is a singleton set if it consists of one point.
This singleton property turns out to be crucial for convergence.




We need to determine the local maxima of $F(\tau)$ and so need to
characterize its gradient. The one-sided directional derivative of
$F(\tau)$ in direction $d$ (so that $d$ is a unit vector) is defined by
\begin{displaymath}
  \mathrm{DF}(\tau; d) = \lim_{\alpha \rightarrow 0^{+}} \frac{1}{\alpha}
  \big( F(\tau + \alpha d) - F(\tau) \big).
\end{displaymath}
Then we have the following characterization and a well-known optimality
condition.
\begin{lemma}[Lemma~6 \cite{Selim1984}]
  \label{lem:DF-tau-star-d}
  $\mathrm{DF}(\tau^*; d)$ exists for any $d$ at any point $\tau^*$, and
  is given by
  \begin{equation}
    \label{eq:DF-expr}
    \mathrm{DF}(\tau^*; d) =
    \underset{\Theta \in M(\tau^*)}{\max}
    \left. d^T\frac{\partial \mathcal{L}}{\partial \tau} \right|_{\tau = \tau^*}
    = \underset{\Theta \in M(\tau^*)}{\max}\: \mathcal{L}(d, \Theta).
  \end{equation}
\end{lemma}
\begin{proof}
  See Appendix~\ref{appdix:proof-DF-tau-star-d}.
\end{proof}

\begin{lemma}[\cite{Selim1984,Magnus2019}]
  \label{lem:tau-star-local-max-kmlec}
  $\tau^*$ is a local maximum for k-MLE-c problem if and only if
  $\mathrm{DF}(\tau^*; d) \leq 0$  for all feasible $d$.
\end{lemma}


\begin{proposition}[\cite{Selim1984}]
  \label{thm:local-maximum-kmlec}
  Given $(\tau^*,\Theta^*)$ where $\tau^* \in \mathcal{B}$ and
  $\Theta^* \in M(\tau^*)$, $\tau^*$ is a local maximum of k-MLE-c
  problem if and only if
  \begin{equation}
    \label{eq:F-geq-max-L}
      F(\tau^*) = \mathcal{L}(\tau^*, \Theta^*)
      \geq \underset{\tau \in \mathcal{A}}{\max}\; \mathcal{L}(\tau,
      \Theta) \quad \text{for all } \Theta \in M(\tau^*).
  \end{equation}
\end{proposition}
\begin{proof}
  See Appendix~\ref{appdix:proof-thm-local-maximum-kmlec}.
\end{proof}


Now we are ready to present our main result on the local maximum of k-MLE
problem.

\begin{theorem} 
  \label{thm:conv-singleton-local-maxima}
  Let $(\tau^*, \Theta^*)$ be a partial maximum for k-MLE problem.
  Suppose $M(\tau^*)$ is a singleton set, then $\tau^*$ is a local maximum
  of k-MLE problem.
\end{theorem}
\begin{proof}
  See Appendix~\ref{appdix:proof-thm-conv-singleton-local-maxima}.
\end{proof}

\subsection{Convergence under MLE Uniqueness} 
\label{subsec:conv-uniq-mle}

To establish the singleton property of $M(\tau^*)$, we need conditions
to ensure the uniqueness of MLE. %
We first introduce the following classic result.

\begin{theorem}[Corollary~2.5 \cite{Makelainen1981}]
  \label{thm:mle-uniq-cond}
  Suppose that the log-likelihood function $\lam(\theta)$ obeys the
  log-likelihood uniqueness conditions:
  \begin{enumerate}[label=(\roman*)]
  \item ${\lam}(\theta)$ is twice continuously differentiable for
    $\theta \in \Omega$, where $\Omega$ is a connected open subset with
    boundary $\partial \Omega$.

  \item $\lim_{\theta \rightarrow \partial \Omega}
    {\lam}(\theta) = c$ where $-\infty \leq c < \infty$.

  \item the Hessian matrix
    $H = \big\{ \frac{\partial^2 {\lam}}{\partial \theta_i \partial
      \theta_j} \big\}$ is negative definite at every stationary point
    of the likelihood, i.e.
    $\frac{\partial {\lam}}{\partial \theta} = 0$.
  \end{enumerate}
  Then
  \begin{itemize}
  \item There is a unique maximum likelihood estimate
    $\hat{\theta} \in \Omega$.

  \item The log-likelihood function attains: no other maxima in $\Omega$; no
    minima or other stationary points in $\Omega$; its infimum value $c$ on
    $\partial \Omega$ and nowhere else.
  \end{itemize}
\end{theorem}
{\it Remark}. We state the result for the log-likelihood whereas 
\cite{Makelainen1981} state it for the likelihood. But the statements
are equivalent since the Hessians at a stationary point are equivalent.

\begin{proposition}
  \label{prop:cond-Mtau-singleton}
  Suppose that each cluster-specific log-likelihood satisfies the conditions of
  Theorem~\ref{thm:mle-uniq-cond}. Then $M(\tau)$ is a
  singleton set for all $\tau$.
\end{proposition}
\begin{proof}
  We denote $\mathrm{dim}(C_k) = n_k$; the $x_n \in C_k$ as $x_{[n]}$,
  $1 \leq n \leq n_k$;
  $X_1^{n_k} = \big( x_{[1]}, \dots, x_{[n_k]} \big)$.
  In view of the representation of $M(\tau)$ as a Cartesian product, we have only to
  show that $M_k(\tau_k)$ is a singleton for each $k$. This will follow
  if we show that the MLE of $\theta_k$ from the joint density
  $\pi(X_1^{n_k}; \theta_k) = \prod_{n\in C_k} p(x_{[n]}; \theta_k) =
  \prod_{n = 1}^{n_k} p(x_{[n]}; \theta_k)$ is unique. This
  now follows from Theorem~\ref{thm:mle-uniq-cond}.
\end{proof}
{\it Remark}.
In Proposition 10 rather than applying the conditions of Theorem 9 to establish MLE uniqueness,
it is sometimes possible to establish MLE uniqueness directly or by other means. 
We will do this for k-VARs below.

We can now state the main result.

\begin{theorem}
 \label{thm:convergency-min}
  Suppose that each cluster-specific likelihood satisfies the uniqueness conditions
  of Theorem~\ref{thm:mle-uniq-cond}.
  Then, the k-MLE  iterates converge to a local maximum
  point for the k-MLE problem.
\end{theorem}

\begin{proof}
  By Theorem~\ref{thm:conv-parti-max-finite} the k-MLE algorithm
  iterates converge to a partial maximizer for the k-MLE problem. The
  Theorem~\ref{thm:convergency-min} uniqueness conditions guarantee via
  Proposition~\ref{prop:cond-Mtau-singleton} that each $M_k(\tau_k)$ is
  singleton. So by Theorem~\ref{thm:conv-singleton-local-maxima} the
  iterates converge to a local maximum for the k-MLE problem, completing
  the proof.
\end{proof}
{\it Remark}.
As indicated in the previous remark we can alternatively directly show
MLE uniqueness when possible.

\subsection{Convergence of k-Bregman}
\label{subsec:appl-kmle-converg-kbreg}

%
We could now apply Proposition~\ref{prop:cond-Mtau-singleton},
Theorem~\ref{thm:convergency-min} to k-Bregman. However it is simpler
instead to replace Theorem~\ref{thm:convergency-min} with yet another
result from \cite{Makelainen1981}.

\begin{theorem}[Theorem~2.6 \cite{Makelainen1981}]
  \label{thm:mle-uniq-cond-breg}
  Suppose the log-likelihood function $\lam(\theta)$ obeys
  the log-likelihood uniqueness conditions (i),(ii),(iii).
  \begin{enumerate}[label=(\roman*)]
  \item $\lam(\theta)$ is twice continuously differentiable
    for $\theta \in \Omega$ a connected open subset with boundary
    $\partial \Omega$.

  \item The gradient $\frac{\partial \lam}{\partial \theta}$
    vanishes for at least one point $\theta \in \Omega$.

  \item the Hessian matrix
    $H = \big\{ \frac{\partial^2 \lam}{\partial \theta_i
      \partial \theta_j} \big\}$ is negative definite at every point
    $\theta \in \Omega$.
  \end{enumerate}
  Then
  \begin{enumerate}[label=(\alph*)]
  \item $\lam(\theta)$ is concave in $\theta$.

  \item There is a unique maximum likelihood estimate
    $\hat{\theta} \in \Omega$.

  \item The log-likelihood has no other maxima, minima or stationary
    points in $\Omega$.
  \end{enumerate}
\end{theorem}

So we have to check the Theorem~\ref{thm:mle-uniq-cond-breg} conditions.
In view of \eqref{eq:bregman-prob} we need only check
Theorem~\ref{thm:mle-uniq-cond-breg} for exponential family densities.
The density of an exponential family $\ln p(x; \theta)$ in
\eqref{eq:bregman-prob} has the equivalent form
\cite[Section~4]{Banerjee2005}
\begin{equation*}
  \ln p(x; \theta) = \theta^T x - \psi(\theta) - \ln p_0(x)
\end{equation*}
where $\psi(\theta)$ is the log partition function
\begin{equation*}
  \psi(\theta) = \ln Z(\theta) = \ln \int p_0(x)
  \exp({\theta^T x}) \mathrm{d}x
\end{equation*}
and is convex \cite{Banerjee2005}, and $p_0(\cdot)$ is an arbitrary
function endowing a measure (see \cite[Section~4.1]{Banerjee2005}). This means
that the log-density is concave and so the Hessian is negative-semi
definite at each point of $\Omega$. But we need negative definiteness
and to see what that entails we need to explicitly calculate the
Hessian.

We find the score function
 $ \frac{\partial \ln p(x;\theta)}{\partial \theta} =
  x - \frac{\partial \psi}{\partial \theta}$
and
\begin{equation*}
  \frac{\partial \psi}{\partial \theta}
  = \int \frac{x e^{\theta^T x}}{Z(\theta)} \mathrm{d}x
  = \int x p(x;\theta) \mathrm{d}x
  = \mathbb{E}(X) \triangleq \mu(\theta),
\end{equation*}
hence
\begin{math}
  \displaystyle \frac{\partial \ln p}{\partial \theta}
  = x - \mu(\theta).
\end{math}
Then we have
\begin{equation*}
  \begin{array}{r@{\;}l}
    \displaystyle
    - \frac{\partial^2 \ln p}{\partial \theta \partial \theta^T}
    &= \displaystyle\frac{\partial^2 \psi}{\partial \theta \partial \theta^T}
      = \int x \frac{\partial p}{\partial \theta} \mathrm{d}x
      = \int x \frac{\partial \ln p}{\partial \theta^T} p\:
      \mathrm{d}x \\[8pt]
    &\displaystyle = \int x (x-\mu)^T p \mathrm{d}x
      = \int (x-\mu) (x-\mu)^T p \mathrm{d}x \\[8pt]
    &\displaystyle = \mathrm{Var}(X) = \Sigma(\theta),
  \end{array}
\end{equation*}
in which $\mathbb{E}(X)$ and $\mathrm{Var}(X)$ denote the mean and
variance of random variable $X$.
Now we can state the Bregman result.
\begin{proposition}
  \label{prop:convergency-min-breg}
  Suppose the cluster-specific exponential families each have: at least
  one point $\theta_k \in \Omega$ where the gradient of the
  log-likelihood vanishes; positive definite variance matrices
  $\Sigma(\theta_k)$. Then, the k-MLE/k-Bregman iterates converge to a
  local minimum point for the k-MLE/k-Bregman problem.
\end{proposition}
\begin{proof}
  The proof is the same as that of Theorem~\ref{thm:convergency-min} we
  just have to check the conditions of
  Theorem~\ref{thm:mle-uniq-cond-breg}. Following the discussion above,
  we find that the cluster specific Hessian is
  \begin{equation*}
    - \textstyle\sum_{n=1}^{n_k} \Sigma(\theta_k) = -n_k \Sigma(\theta_k)
  \end{equation*}
  which is negative definite as required for condition (iii) in
  Theorem~\ref{thm:mle-uniq-cond-breg}. Condition (i) is implicit for
  exponential families. Condition (ii) holds by assumption. The result
  now follows.
\end{proof}

%



\section{Clustering Autocorrelated Time Series with \lowercase{k}-VARs}
\label{sec:ts-clustering}

We apply k-MLE to clustering of $N$ autocorrelated time series 
$\mathbb{Y} \triangleq\{ Y^{(n)},n=1,2,\cdots,N\}$
where each $Y^{(n)}$ contains $T$ measurements of an $m$-dimensional time series
$Y_t^{(n)}, t=1,2,\cdots,T$.

\subsection{k-VAR\lowercase{s} Derivation}
\label{sec:ts-derive}


We model a time series in the $k^{th}$ cluster 
by a Gaussian
 vector autoregression (VAR), of block order $p_k$.
%
%
The conditional likelihood function of the $k^{th}$
VAR model is
\begin{equation}
  \label{eq:lld-nk}
  \begin{array}{@{}r@{\;}l}
    L_{n,k} &= p(Y^{(n)} \mid A_k, \Sg_k) \\
            &= \displaystyle\prod_{t=p_k+1}^T
              (2\pi)^{-\frac{m}{2}} |\Sg_k|^{-\frac{1}{2}}
              \exp\left( -\frac{1}{2} \mathbf{e}_{nkt}^T \Sg_k^{-1}
              \mathbf{e}_{nkt}\right)
  \end{array}
\end{equation}
where for each $k$
\begin{align*}
  \mathbf{e}_{nkt} &= Y_t^{(n)} - A_{k0} - A_{k1}Y_{t-1}^{(n)} -
                    -\cdots- A_{kp_k}Y_{t-p_k}^{(n)} \\
                   &= Y_t^{(n)} - A_k X_{kt}^{(n)}, \\
  A_k &= \left[A_{k0}, A_{k1}, \dots, A_{kp_k} \right],\\
  X_{kt}^{(n)} &= \left[ {1}, Y_{t-1}^{(n)T}, Y_{t-2}^{(n)T}, \dots,
                 Y_{t-p_k}^{(n)T} \right]^T,
\end{align*}
Also $A_{k0}$ is a vector, while for $r\geq 1$, $A_{k,r}$ are the VAR coefficient matrices and
$\Sg_k$ the driving noise variance matrix.
%
The set-up now fits in the k-MLE framework.

The classification log-likelihood is given by
\EBN
  \mathcal{L}(\tau,\Theta) = \textstyle\sum_{n=1}^N \sum_{k=1}^K
  \tau_{n,k} \lam_{n,k}(\theta_k)                 \label{eq:slik}
\EEN
where 
$\theta_k$
denotes the collection of parameters $A_k$ and $\Sg_k$.
Applying the coordinate ascent solver for k-MLE
is straightforward, since we have a classic vector or multivariate regression which
leads to the following
algorithm:
\begin{itemize}
\item label update:
  \begin{align}
    \label{eq:E-kAR-tau}
    \tau_{n,k} &=
    \begin{cases}
      1 & \text{if } k = \argmin_{k'} D_{n,k'}, \\
      0 & \text{otherwise},
    \end{cases}\\
    \label{eq:E-kAR-tau-Dnk}
    D_{n,k'} &= (T-p)\log |\Sg_{k'}| + \textstyle\sum_{t=p+1}^T \mathbf{e}_{nk't}^T
              \Sg_{k'}^{-1} \mathbf{e}_{nk't},
  \end{align}

\item parameter update:
  \begin{align}
    \label{eq:E-kAR-Ik}
    I_k &= \big\{n: \tau_{n,k} = 1;\ n = 1,\dots,N \big\}, \\
    \label{eq:M-kAR-theta}
    A_k^T &= \textstyle
       \begin{array}[t]{@{}l@{}l}
         &\Big( \sum_{n \in I_k} \big( \sum_{t=p+1}^T X_{kt}^{(n)}
           X_{kt}^{(n)T}  \big) \Big)^{-1} \\
         &\Big( \sum_{n \in I_k} \big( \sum_{t=p+1}^T
           X_{kt}^{(n)}Y_{t}^{(n)T} \big) \Big),
       \end{array} \\
    \label{eq:M-kAR-omega}
    \Sg_k &=
       \Big(\textstyle\sum_{n \in I_k}
            \big( \sum_{t=p+1}^T \mathbf{e}_{nkt} \mathbf{e}_{nkt}^T \big)
       \Big) /
       \Big( (T-p) |I_k| \Big),
  \end{align}
  where $|I_k|$ denotes the number of elements of $I_k$ (i.e. the
  cardinality of set $I_k$), and $p \triangleq \max \{p_1, \dots, p_K\}$.
\end{itemize}

We now consider:
initialisation,
stopping conditions, and fast matrix computation. 

\subsection{k-VARs Computation}
\label{sec:kARs-computation}


\subsubsection{Initialisation}
\label{subsec:initialisation}

We take a simple approach.
We fit a VAR model to each
 time series $Y^{(n)}$, yielding
$(A_n^*, \Sg_n^*)$ ($n=1,\dots,N$). 
Then we choose at
    random, one $A_n^*$ (and $\Sigma_n^*$) to initialise the
    cluster-specific parameters.  
We call k-VARs with this initialisation `k-VARs(rnd)'.
   In practice we can further repeat this many times and choose 
   the initialisation that delivers the highest value of the likelihood to
    improve performance. 
    If these $K$ initial parameters are all from different clusters 
    then we call the result ‘k-VARs (oracle)’.


%

\subsubsection{Stopping Criteria}
\label{subsec:stopping-criterion}
We considered  two stopping criteria:
one based on parameters, the other based on log-likelihoods.

If $A_k$ and $\Sg_k$ are the values
given at the current iterate and $A^+_k$ and $\Sg^+_k$
are the next, the stopping condition is
\begin{equation}
  \label{eq:stop-condition}
  \max\left\{ \|A_k^+ -
    A_k\|_F,  \|\Sg_k^+ - \Sg_k\|_F \right\} < \epsilon
\end{equation}
for all $k = 1,\dots,K$, where $\epsilon$ is the user-specified tolerance
value. One may also consider using the relative errors in
\eqref{eq:stop-condition} according to practical demands.

The log-likelihood stopping rule is
\begin{equation}
  \label{eq:stop-condition-fast}
  |\mathcal{L}(\tau^+, \theta^+) - \mathcal{L}(\tau, \theta)| < \epsilon,
\end{equation}
where $\tau^+$ and $\theta^+$ denote the proposed updates. 
Compared to the parameter stopping rule, this one is computationally
much cheaper since it avoids computing large matrix norms.

\subsubsection{Fast Computation}
\label{subsec:fast-computation}

The computation bottleneck of the k-VARs algorithm is
\eqref{eq:M-kAR-theta} and \eqref{eq:M-kAR-omega}. To accelerate
computation, we use QR decomposition as now described.

We introduce extended vectors of the general form, $\Fbfn\T=[F_{p+1},\cdots,F_T]$ 
and corresponding outer products of the form, $\Fbfn\Gbfn\T=\summb{p+1}{T}F_tG_t\T$.
We thus define $\Xbfnk,\Ybfn,\Ebfnk=\Ybfn-\Xbfnk A_k\T$ and
$\Xbfnk\T\Xbfnk,\Xbfnk\T\Ybfn,\Ebfnk\T\Ebfnk$.

We now introduce the QR decompositions
\begin{equation}
[\Xbfnk]_{(T-p)\times(1+m\pk)}=(\Qnk)_{(T-p)\times(1+m\pk)}\Rnk \nonumber
\end{equation}
where $\Qnk$ is orthogonal and $\Rnk$ is upper triangular.
We then form $\Ybfqnk=\Qnk\T\Ybfn$ which 
yields a new expression for
\eqref{eq:M-kAR-theta} as follows
%
%
\begin{equation}
  \label{eq:M-kAR-theta-fast}
  A_k^T =
  \left( \textstyle\sum_{n \in I_k} R_{nk}^T R_{nk} \right)^{-1}
  \left( \textstyle\sum_{n \in I_k} R_{nk}^T \mathbf{Y}_{Q_{nk}} \right).
\end{equation}
The QR decomposition also enables cheap computation of individual
time series parameters as
\begin{equation}
  \label{eq:VAR-theta}
  A_n^{*T} = R_{nk}^{-1} \mathbf{Y}_{Q_{nk}}.
\end{equation}

We next introduce a second set of QR decompositions
\begin{equation}
  \label{eq:QR-mat-E}
  \Ebfnk=(U_{nk})_{(T-p)\times m}(V_{nk})_{m\times m}
\end{equation}
This yields a compact  update of $\Sg_k$ as follows
\begin{equation}
  \label{eq:M-kAR-omega-fast}
  \Sg_k = \left( \textstyle\sum_{n \in I_k} V_{nk}^T V_{nk} \right)
  / \big( (T-p) |I_k| \big),
\end{equation}
The initial covariance for each time series
can be similarly computed by
\begin{equation}
  \label{eq:VAR-omega}
  \Sg_n^* = \big( 1/(T-p_k) \big) V_{nk}^T V_{nk}.
\end{equation}

Finally we introduce the lower triangular Cholesky decompositions
\begin{math}
  \Sg_k = L_k L_k^T, \, k = 1,\dots,K,
\end{math}
This yields reliable computation of $D_{nk}$ as follows.
\begin{equation}
  \label{eq:E-kAR-psi-fast}
  \mathbf{e}_{nkt}^T \Sg_k^{-1} \mathbf{e}_{nkt} =
  (L_k^{-1} \mathbf{e}_{nkt})^T (L_k^{-1} \mathbf{e}_{nkt}),
\end{equation}
where $L_k^{-1} \mathbf{e}_{nkt}$ is found by rapidly solving the triangular linear
system $L_k \xi = \mathbf{e}_{nkt}$ for $\xi$.
which is very fast since $L_k$ is lower triangular.


\subsubsection{Summary of k-VARs and Parallelism} 
\label{subsec:parallel-computation}

The complete k-VARs algorithm is summarised in Algorithm~\ref{alg:k-ARs}.
In addition, it is easy to see that several
sections of Algorithm~\ref{alg:k-ARs} can be parallelised to take
advantage of multi-cores, including:
\begin{itemize}
\item (line 3) pre-computing $n$, QR decompositions and $K$
  estimations of $A_n^*$.
\item (line 6-11) computing $NK$, $D_{n,k}$s.
\item (line 12-16) updating $K$ matrix parameters
  $A_k, \Sigma_k$.
\end{itemize}

\begin{algorithm}
  \caption{k-VARs algorithm}
  \label{alg:k-ARs}
  \begin{algorithmic}[1]
    \State \textsc{Input}: data $\mathbb{Y}$, number of clusters $K$,
    model orders $p_1,\dots,p_K$, and tolerance $\epsilon$.

    \State \textsc{Output}: model parameters
    $A_k, \Sg_k$, and clustering labels
    $\tau_{n,k}$ ($k = 1,\dots,K$; $n=1,\dots,N$).

    \State Pre-computation: perform and cache QR decomposition
    for each $\mathbf{X}_{nk}$, and compute
    $\mathbf{Y}_{Q_{nk}}$.

    \State \textsc{Initialisation}: initialise
    $(A_k, \Sg_k)$ by Section~\ref{subsec:initialisation}.


    \While{TRUE}

    \For{$n \gets 1$ to $N$}\Comment{label update}
    \For{$k \gets 1$ to $K$}
    \State Compute $D_{n,k}$ by \eqref{eq:E-kAR-tau-Dnk}.
    \EndFor
    \State Determine $\tau_{n,k}$ by \eqref{eq:E-kAR-tau} for $k=1,\dots,K$.
    \EndFor

    \For{$k \gets 1$ to $K$}\Comment{parameter update}
    \State Determine $I_k$ by \eqref{eq:E-kAR-Ik}.
    \State Update $A_k$ by \eqref{eq:M-kAR-theta-fast}.
    \State Update $\Sg_k$ by \eqref{eq:M-kAR-omega}.
    \EndFor

    \If{\eqref{eq:stop-condition} or \eqref{eq:stop-condition-fast}
      satisfied}\Comment{stop conditions}
    \State \textbf{break} \EndIf

    \EndWhile
  \end{algorithmic}
\end{algorithm}



\subsection{k-VARs Convergence}
\label{sec:convergence-analysis-kARs}

Convergence analysis of k-VARs algorithm is provided by applying the
general k-MLE results. 
As indicated in the remarks following  Proposition 9 and Theorem 10
we will directly show the solution to the MLE equations is unique.

Note that the model is not from a (vector) exponential family.
This is because the exponent can only be represented as an inner product of parameters
with data if $\Sg_k$ is known. So we cannot apply (a vector version of) k-Bregman.

Differentiating a cluster-specific likelihood gives the least squares equations shown in (21),(22).
From these it is implicit that uniqueness requires that, for each $k$:
(i) the inverse of the $\Xbf_{nk}^T\Xbf_{nk}$ matrix in (21) exists; 
(ii) $\Sg_k$ in (21) has full rank.
This will follow if (i),(ii) hold for every cluster of size 1.
But we are now in a classic situation of requiring uniqueness
of the least squares estimate of a VAR and of its noise variance from a single time series record.
For cluster size 1,  the $\Xbf_{nk}^T\Xbf_{nk}$ matrix in (22) is very nearly a block Toeplitz matrix.
So long as each time series is generated by 
(i) a \ul{stable} VAR with
(ii) full rank noise covariance matrix
and $T>p+1$,
then the $\Xbf_{nk}^T\Xbf_{nk}$ matrix will have full rank with probability 1 
\cite{Lutkepohl2005}. Further the estimated noise variance matrix will
have full rank with probability 1 \cite{Lutkepohl2005}.
So uniqueness of the MLE is established and convergence of the
k-VARs algorithm now follows.

\subsection{Model Selection for k-VARs}
\label{sec:model-selection-kARs}

The clustering algorithms, presented in section~\ref{sec:ts-derive}
and \ref{sec:kARs-computation}, require the number of clusters
$K$ and cluster-specific model orders $(p_1, \dots, p_K)$ to be
specified. This section develops a \emph{Bayesian
  information criterion} (BIC) to resolve the model selection problem. 
We only discuss the
special case with equal $p_k$'s.
The general case is  computationally challenging
and may not offer big improvements on large data sets.

%
%


For the case of equal $p_k$'s,
 we choose $K$ and $p$ as the joint minimizer
 of BIC on a $(K,p)$ grid as follows.
\begin{equation}
  \label{eq:BIC-K-p}
  \begin{array}{@{}r@{\;}l}
    \mathrm{BIC}(K,p) =
    &- 2 \log {L} +
      \Big\{ K \big[ (p+ 1/2) m^2 + 3m/2 \big] + N \Big\} \cdot\\
    & \log\big[ N(T-p) \big] ,
  \end{array}
\end{equation}
Since the $N$
time series are assumed to be sampled independently and the
$K$ models are independent then the likelihood
is given by (\ref{eq:slik})
%
%
with all $p_k$ in $D_{n,k}(\cdot, \cdot)$ being set to $p$.

A faster way to find the minimum of BIC is to use cyclic descent.
Here we alternate between the following two steps until a convergence criterion is met.\\
\mbh (i) Given $p$, minimize over $K$;\\
\mbh (ii) Given $K$, minimize over $p$.\\
Of course this may get stuck in a local minimum.
But that could be managed by trying a number of random starts.


\section{Simulation and Application  to Real Data}
\label{sec:experiments}

This section has two parts.
In part I (subsections A,B,C) we
we compare k-VARs with other methods on simulated data.
We find that k-VARs considerably outperforms 
state-of-the-art methods
because they ignore the autocorrelation feature.
Section C studies the robustness of k-VARs with non-Gaussian driving noise
and with varying signal to noise ratios.
In part II (section D) we apply k-VARs to real data.


\subsection{Baseline Methods and Performance Indices}
\label{sec:basel-meth-perf}

%
k-VARs is implemented in MATLAB. 
%
All simulations are executed on 2.4 GHz Intel Core i9 machines
with 32GB RAM and macOS Catalina. The  `state of the art'
comparator methods are as follows.
\begin{itemize}
\item k-DBA: k-Means with DTW barycenter averaging (DBA)
  \cite{Petitjean2011}, the implementation \texttt{TimeSeriesKMeans}
  with option \texttt{metric="dtw"} from an official Python package
  \cite{tslearn}, \texttt{tslearn}, version 0.5.0.5.

\item k-Shape: the well-known algorithm proposed in
  \cite{Paparrizos2015,Paparrizos2017}, the official implementation
  \texttt{KShape} from the same Python package \texttt{tslearn}.

\item k-GAK: Kernel k-Means \cite{Dhillon2004} with Global Alignment
  Kernel \cite{Cuturi2011} for time series clustering, the
  implementation \texttt{KernelKMeans} with option \texttt{kernel="gak"}
  from the Python package \texttt{tslearn}.

\item k-SC: k-Spectral Centroid proposed in \cite{Yang2011}, the MATLAB
  implementation given on the companion website of SNAP software
  \cite{Leskovec2014}. (Note: this version does not include the
  Haar-Wavelet based incremental implementation.)

\end{itemize}

To avoid the drawbacks of Rand Index (RI) and the squared version of
Normalised Mutual Information ($\text{NMI}_{\text{sqrt}}$) discussed
in \cite{VEPP10},
we used the improved measures, adjusted Rand Index (ARI)
and Normalized Information Distance (NID) to evaluate clustering
performance
\begin{align*}
  &\mathrm{ARI} = \frac{2(N_{00}N_{11} - N_{01}N_{10})}{
                 (N_{00} + N_{01})(N_{01} + N_{11}) +
                 (N_{00} + N_{10})(N_{10} + N_{11})}, \\
  &\mathrm{NMI}_{\text{max}} = \frac{I(U,V)}{\max\{H(U), H(V)\}},
  \quad \mathrm{NID} = 1 - \mathrm{NMI}_{\text{max}}
\end{align*}
where $N_{11}$ is the number of pairs that are in the same cluster in
both $U$ and $V$, $N_{00}$ is the number of pairs that are in different
clusters in both $U$ and $V$, $N_{01}$ is the number of pairs that are
in the same cluster in $U$ but in different clusters in $V$, $N_{10}$ is
the number of pairs that are in different clusters in $U$ but in the
same cluster in $V$; and $H(U)$ is the average amount of information in
$U$, $I(U,V)$ is the mutual information, given as
\begin{equation*}
  \begin{array}{r@{\;}l}
    H(U) &= - \sum_{i=1}^K \frac{a_i}{N} \log \frac{a_i}{N},\\
    I(U,V) &= \sum_{i=1}^K \sum_{j=1}^K \frac{n_{ij}}{N}
             \log \frac{n_{ij}/N}{a_i b_j / N^2},
  \end{array}
\end{equation*}
in which $n_{ij}$ denotes the number of objects that are common to
clusters $U_i$ and $V_j$, $a_i$ is the number of objects in cluster
$U_i$, and $b_j$ is that in cluster $V_j$. For both ARI and
$\mathrm{NMI}_{\text{max}}$, values close to 1 indicate high clustering
performance.

\subsection{Comparisons on Simulated Datasets} 
\label{subsec:synthetic-datasets}

To evaluate performance in a statistical manner, we randomly generate
$40$ independent datasets. Due to scalability limits on some of the competing methods,
the problem size had to be restricted.
%
Each time-series
dataset is generated by simulating a class of VAR models with the
following specifications:
\begin{itemize}\setlength\itemsep{-.2em}
\item time series dimensions: $m = 2, 4, 8$;
\item model order: $p = 5$;
\item time series length: $T = 80$;
\item number of groups: $K = 8$;
\item number of time series per cluster: $N_c = 30$.
\end{itemize}
The parameters of each VAR model are generated randomly and 
roots scaled to ensure stability. 
The noise covariance matrix is computed from randomly generated Cholesky factors.
%

\begin{figure}[htbp]
  \centering

  \begin{subfigure}[b]{0.5\textwidth}
    \centering
    \includegraphics[width=\textwidth]{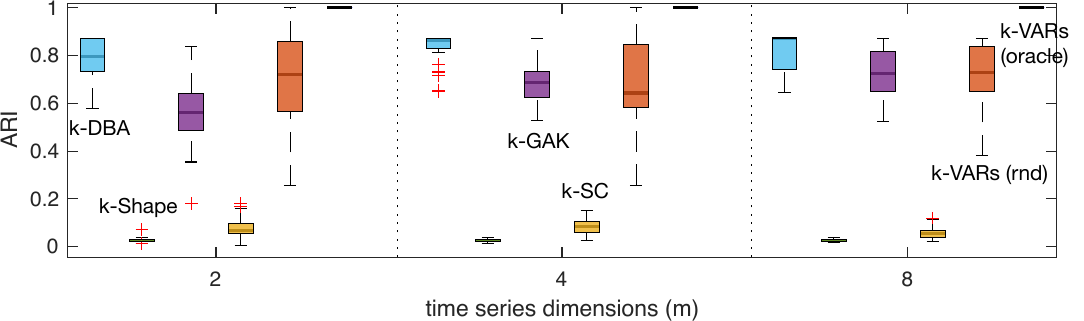}
    \caption{Adjusted Rand Index (ARI)}
    \label{subfig:boxplot-RI}
  \end{subfigure}
  \\
  \begin{subfigure}[b]{0.5\textwidth}
    \centering
    \includegraphics[width=\textwidth]{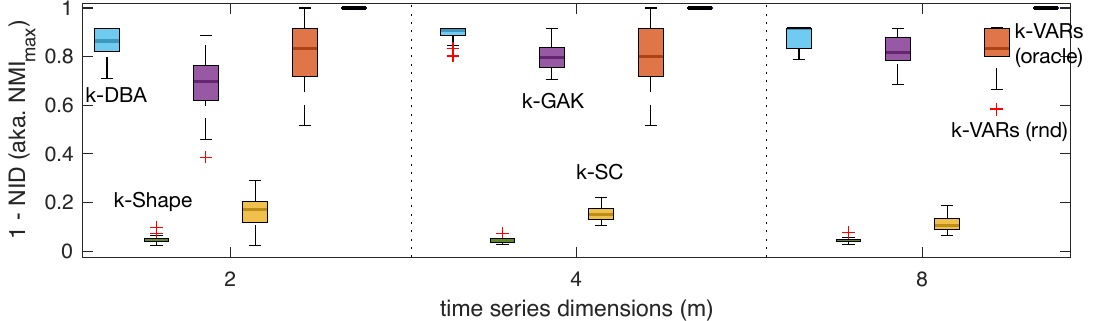}
    \caption{Normalised Information Distance (NID)}
    \label{subfig:boxplot-NMI}
  \end{subfigure}

 \caption{Comparison of clustering performance of k-VARs with
    state-of-the-art methods.`k-VARs (rnd)'
     initializes each cluster with a randomly chosen time series.
    'k-VARs (oracle) initialises each cluster with a randomly
    chosen time series from the true cluster.}
  \label{fig:bmk-clust-prec}

\end{figure}

The  results in
Figure~\ref{fig:bmk-clust-prec}, show two versions of k-VARs (defined in the caption).
k-VARs (oracle), is superior to the state-of-the-art methods. 
And k-VARs (rnd) is 
competitive with them.
%
%
Note also that most of the other methods require a number of tuning parameters
to be chosen.
We used default values suggested in the various implementations.
k-VARs has no free tuning parameters and is competitive
because the other methods ignore
the autocorrelation feature.

\begin{figure}[htbp]   
  \centering
  \includegraphics[width=.48\textwidth]{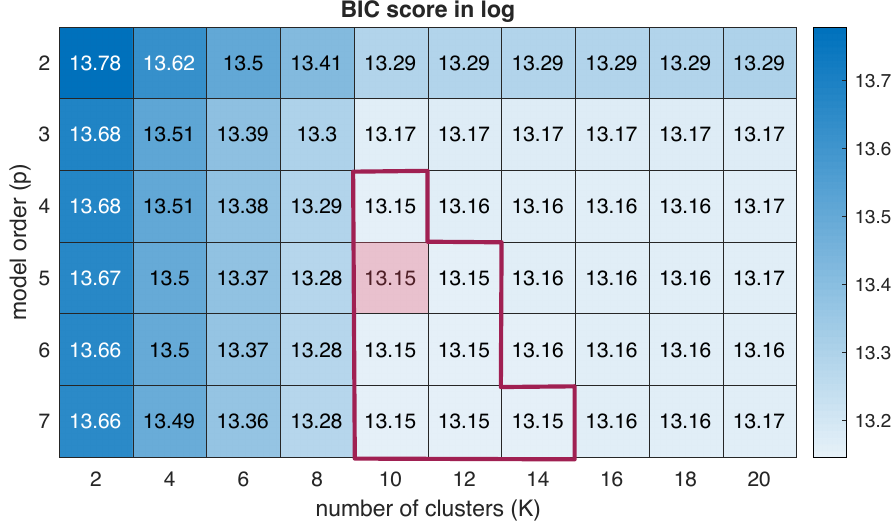}
  \caption{Log-scale heatmap of BIC scores as a function of number of
    clusters $K$ and model order $p$. The minimum
     is 13.15 (surrounded by red lines),
    attained at $K = 10$ and $p = 4$ (the ground truth is
    filled in red). 
    }
  \label{fig:BIC-mesh}
\end{figure}

Next we illustrate BIC.
The following set-up applied.
\begin{itemize}
\item cluster-specific VAR models: $m = 4$, $p = 5$;
\item time series: $T = 200$, $K = 10$, $N_c = 20$.
\end{itemize}
The grid for $K,p$ is $2\!:\!2\!:\!20\times 2\!:\!1\!:\!8$. 
log-BIC is plotted as a heatmap in
Figure~\ref{fig:BIC-mesh}.

We see that BIC attains its minimum 
at the ground
truth $K=10$.  However there is a relatively flat region in the vicinity of the minimum.
We found in the simulations
that  good clustering performance
was  tolerant to some flatness in BIC near its minimum.
%

\subsection{Effect of Non-Gaussian Driving Noise and of SNR} 
Here we study two things. 
The ability of k-VARS to deal with non-Gaussian driving noise.
And the effect of signal to noise ratio (SNR).

\begin{figure}[htbp]
  \centering
  \begin{subfigure}[b]{0.5\textwidth}
    \centering
    \includegraphics[width=\textwidth]{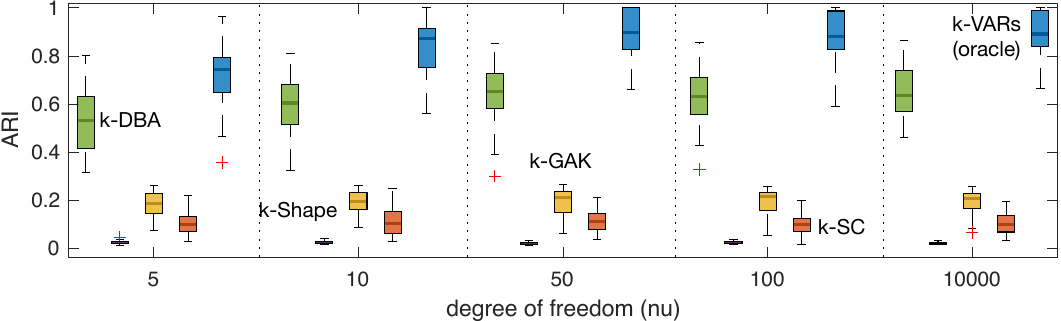}
    \caption{Adjusted Rand Index (ARI)}
    \label{subfig:boxplot-ARI-stut}
  \end{subfigure}
  \\
  \begin{subfigure}[b]{0.5\textwidth}
    \centering
    \includegraphics[width=\textwidth]{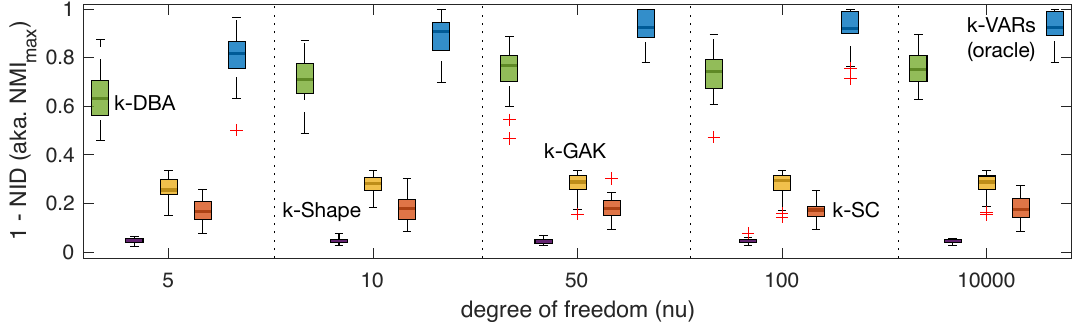}
    \caption{Normalized Information Distance (NID)}
    \label{subfig:boxplot-NMI-stut}
  \end{subfigure}
  \caption{
  Effect of t-distributed driving noise,
  on clustering performance of k-VARs compared to
    state-of-the-art methods.}
  \label{fig:bmk-clust-prec-stut}
\end{figure}

We use t-distributed driving noise
with degrees of freedom (dof): $[5, 10, 50, 100, 10000]$.
Larger dof is closer to Gaussianity.
The setup is otherwise the same as 
for Figure~\ref{fig:bmk-clust-prec}, but with $m=3$. 
The results in
  Figure~\ref{fig:bmk-clust-prec-stut} illustrate the robustness of k-VARs 
 with little loss of performance for low dof.

\begin{figure}[htbp]
  \centering
  \begin{subfigure}[b]{0.5\textwidth}
    \centering
    \includegraphics[width=\textwidth]{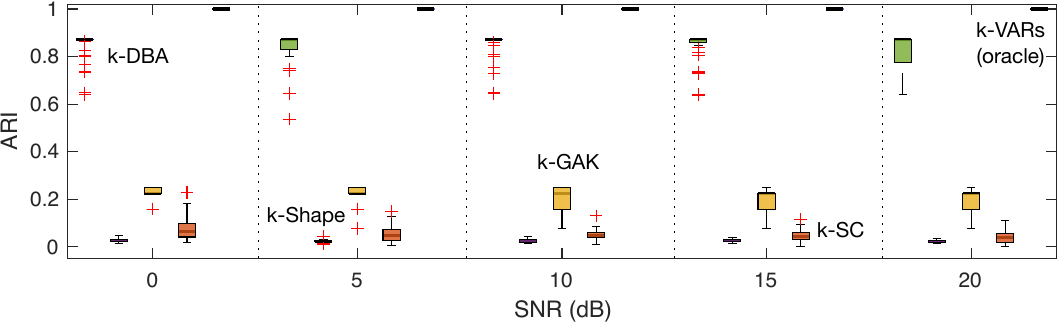}
    \caption{Adjusted Rand Index (ARI)}
    \label{subfig:boxplot-ARI-snr}
  \end{subfigure}
  \\
  \begin{subfigure}[b]{0.5\textwidth}
    \centering
    \includegraphics[width=\textwidth]{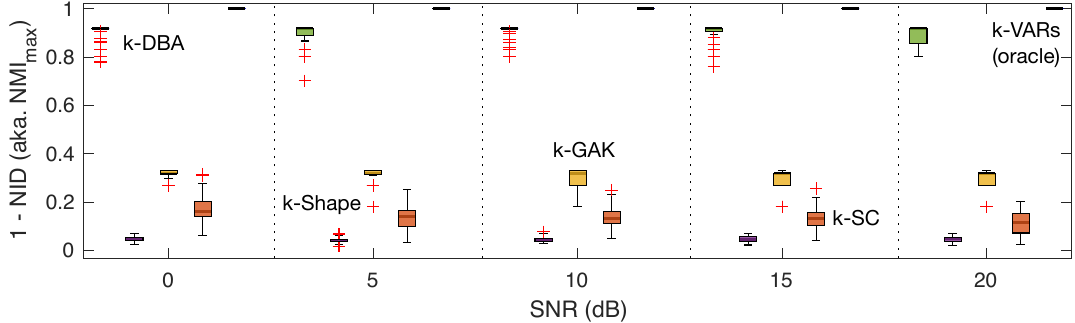}
    \caption{Normalized Information Distance (NID)}
    \label{subfig:boxplot-NMI-snr}
  \end{subfigure}
  \caption{
  Effect of SNR on
  clustering performance of k-VARs compared to
    state-of-the-art methods.}
  \label{fig:bmk-clust-prec-snr}
\end{figure}

The next simulation studies the effect of SNR for Gaussian driving noise.
We modify a definition of SNR due to \cite{NICS20}.
Denote $\Pi$ as the zero-lag auto-covariance of a stationary VAR(p) time series $Y_t$
whose white noise variance is $\Sg$. 
Then \cite{NICS20} uses 
SNR$=10 \log \mbox{VNSR}, 
\mbox{VSNR}=\frac{\lam_{\mathrm{max}}(\Pi)}{\lam_{\mathrm{max}}(\Sg)}$.
This has a scaling problem. If we multiply $Y_t$ by a diagonal
scaling matrix $L\neq cI$ (where $c$ is a constant) then this SNR changes. 
So instead we use 
VSNR$=\half\lam_{max}(\Sg^{-1/2}\Pi\Sg^{-1/2})$ which is unaffected by matrix scaling.
The divisor of $2$ is due to a second adjustment as follows. 
In steady state we can write $Y_t=\hat{Y}_t+\ep_t$ where $\hat{Y}_t$
is the one step ahead predictor and $\ep_t$ is the one step ahead 
prediction error and so has variance $\Sg$. Also it is uncorrelated with 
$\hat{Y}_t$ which thus has variance $M=\Pi-\Sg$.
In the scalar case the when $M=\Sg$ i.e. `signal' and noise variance are equal
this would be taken to yield a SNR of 0 dB. So to reproduce that in the vector case
we divide by $2$.
  The model setup is the same as that for
  Figure~\ref{fig:bmk-clust-prec-stut}, except we fixed $m=3$.
 We have to modify $A_k$ to achieve a given SNR.
 The trick is to scale the  roots  of the VAR polynomial
 $A_{k_1},\cdots,A_{k_p}$. 
 The SNR values are [0,5,10,15,20] (dB), 
  The results  in
  Figure~\ref{fig:bmk-clust-prec-snr}, show k-VARS to be reliable.

\subsection{Application to Real Dataset \emph{WAFER}}
\label{subsec:real-data}

We now analyse the chip frabrication `WAFER' data set \cite{Olszewski2001}
common in time series clustering studies and available at  \cite{MTSdatasets}.
Each data set in the database contains the
measurements recorded by one sensor during the processing of one wafer
by one tool.

 The results in Table~\ref{tbl:wafer}, show comparisons to
  k-DBA, k-Shape, k-GAK, k-SC. 
  The proposed method performs best by all measures. One may
  recall that the performance of k-VARs is subject to initialization and
  hence is restricted by local optimality. Unlike the Monte Carlo study,
  we can now optimise the initialisation.
  Recalling that k-MLE aims to maximise the
  mixture likelihood, we firstly perform multiple runs with random
  initialisations, and then apply a likelihood  threshold
  to evaluate initialisations. The results for k-VARs in
  Table~\ref{tbl:wafer} show a statistical summary of 20 runs
  with random initialisations selected by thresholding the
  log-likelihood at $15.57$, which value is chosen via 10 preliminary trials.
  The performance of k-VARs is so consistent that its 1st, 2nd, 3rd quartile
   are the same, hence only one value is given in
  Table~\ref{tbl:wafer}.

In addition, our methods provide
cluster-specific parameters for further analyses, like model validation,
Although the manufacturing
processes of WAFER is complex and with higher-order dynamics, the
results show that satisfactory clustering does not require precise
modelling of each time series ($p = 10$ is used here). Again the reason
k-VARs performs well here is that the WAFER data shows strong
autocorrelation features  which the other methods ignore.

\begin{table}[htb]
  \caption{Clustering performance for the WAFER data.}
  \label{tbl:wafer}
  \centering
    \begin{small}
      \begin{sc}
        \begin{tabular}{lccccc}
          \toprule
          & k-DBA & k-Shape & k-GAK & k-SC & k-VARs \\
          \midrule
          RI    & 0.6280 & 0.5914 & 0.5008 & 0.8050 & \textbf{0.8120} \\
          $\text{NMI}_{\mathrm{sqrt}}$   & 0.0001 & 0.0001 & 0.0054 & 0.0043 & \textbf{0.0463} \\
          ARI   & 0.0050 & {\bf 0.0321} & 0.0026 &-0.0057 & 0.0165\\
          1-NID & 0.0001 & 0.0032 & 0.0038 & 0.0011 & \textbf{0.0074} \\
          \bottomrule
        \end{tabular}
        \vskip -0.1in
      \end{sc}
    \end{small}
\end{table}


\section{Conclusion}

In this paper we have developed a general purpose
deterministic label clustering method, k-MLE,
that is not based on a distance or divergence measure,
but on likelihood. 
We developed an implementation based on cyclic ascent.
We also developed a model selection procedure based on BIC
and  illustrated it in a special case.

We provided a general theorem on parameter convergence of the cyclic ascent
iterates. In so doing we drew attention
to major flaws in previous clustering algorithm convergence proofs.
They only obtained criterion convergence, mistakenly asserting
that it implied parameter convergence  whereas proving 
parameter convergence is much much harder.

We showed that a previous clustering method based on Bregman divergence,
which we called k-Bregman, is a special case of k-MLE.
Our general convergence result then provided the
first valid proof of parameter convergence of k-Bregman.

We illustrated the usefulness of k-MLE by developing a new
clustering algorithm, k-VARs, for autocorrelated vector time series.
We developed a fast computational procedure and a BIC
model selection criterion for simultaneously choosing
the number of clusters and VAR order.

We only illustrated BIC for k-VARs, but as indicated in the introduction,
it can be constructed for any k-MLE algorithm, since it requires
only a likelihood and a parameter count.

We compared k-VARs with state of the art vector time series
clustering algorithms and found it greatly outperformed them.
This is simply because those algorithms
mostly ignore the autocorrelation feature.
We also showed robust performance 
in the presence of heavy-tailed driving noise
and also with varying SNR.

\section*{Acknowledgments}

This work was supported by a Discovery Grant
DP180102417 from the Australian Research Council.




\appendices
\section{Proofs for Convergence Analysis}
\label{appdix:proof-conv-analys}

\subsection{Proof of Lemma~\ref{lem:convex-Ftau}}
\label{appdix:proof-lem-convex-Ftau}

Let $0 \leq \alpha \leq 1$ then we have
\begin{displaymath}
  \begin{array}{r@{\;}l}
    & F \big( \alpha \tau^a + (1-\alpha) \tau^b \big) \\
    &= \underset{\Theta \in \Omx}{\max}
      \sum_{n=1}^N \sum_{k=1}^K
      \left[ \alpha \tau_{n,k}^a + (1-\alpha) \tau_{n,k}^b \right]
      \ell(x_n, \theta_k) \\
    &= \underset{\Theta \in \Omx}{\max}
      \left[
      \alpha \mathcal{L}(\tau^a, \Theta) + (1-\alpha)
      \mathcal{L}(\tau^b, \Theta)
      \right] \\
    &\leq \alpha\, \underset{\Theta \in \Omx}{\max}\,
      \mathcal{L}(\tau^a, \Theta) +
      (1-\alpha)\, \underset{\Theta \in \Omx}{\max}\,
      \mathcal{L}(\tau^b, \Theta) \\
    &= \alpha F(\tau^a) + (1-\alpha) F(\tau^b)
  \end{array}
\end{displaymath}
as required.

\subsection{Proof of Proposition~\ref{thm:prob-same-sol}}
\label{appdix:proof-thm-prob-same-sol}

Since the concentrated log-likelihood $F(\tau)$ is convex, and since the
constraint set $\mathcal{A}$ is convex, the solution of the k-MLE-c
problem is an extreme point of $\mathcal{A}$ and so lies in
$\mathcal{B}$. Let $\tau^{*}$ be this solution. Let $\Theta^*$ be any
$\Theta\in\Omx$ that achieves $\mathcal{L}(\tau^*, \Theta^*) = F(\tau^*)$. Then
$(\tau^*, \Theta^*)$ also solves the k-MLE problem. The converse also
holds completing the proof.

\subsection{Proof of Theorem~\ref{thm:conv-parti-max-finite}}
\label{appdix:proof-thm-conv-parti-max-finite}

Since $\mathcal{B}, \Omx$ are bounded sets, the iterates
$(\tau^{(m)}, \Theta^{(m)})$ are a bounded (matrix) sequence and so by
the Bolzano-Weierstrass theorem have at least one limit point/matrix.
Next we claim that an extreme point of $\mathcal{A}$ is visited at most
once before the algorithm stops. Suppose that this is not true, i.e.
$\tau^{(a)} = \tau^{(b)}$ for some $a < b$. In view of the cyclic
minimization we must have
$  \mathcal{L}(\tau^{(b)}, \Theta^{(b)}) \geq
  \mathcal{L}(\tau^{(b)}, \Theta^{(a)}) \geq
  \mathcal{L}(\tau^{(a)}, \Theta^{(a)})$.
If either of the inequalities is strict then we get a contradiction due to
the stopping rule and the claim holds. If both inequalities are
equalities we similarly have a contradiction since the algorithm must
already have stopped. It now follows that, since there are a finite
number of extreme points of $\mathcal{A}$, the algorithm will reach a partial
maximum in a finite number of steps.

\subsection{Proof of Lemma~\ref{lem:DF-tau-star-d}}
\label{appdix:proof-DF-tau-star-d}

The proof in \cite{Selim1984} deals with minima and uses slightly
different assumptions. We note that
Assumption~\ref{assump:compact-Om-cont-ell} ensures
$\mathcal{L}(\tau, \Theta)$ as well as
$\frac{\partial \mathcal{L}}{\partial \tau}$ are continuous in
$(\tau,\Theta)$. Then, by \cite{Lasdon1970} (Theorem~1, p.~420), we
get the first equality in \eqref{eq:DF-expr}. By Property~P1, we
further have
\begin{displaymath}
  \mathrm{DF}(\tau^*; d)
  = \underset{\Theta \in M(\tau^*)}{\max}\:
  d^T \left[ \ell(x_n; \theta_k) \right]
  = \underset{\Theta \in M(\tau^*)}{\max}\: \mathcal{L}(d, \Theta).
\end{displaymath}

\subsection{Proof of Proposition~\ref{thm:local-maximum-kmlec}} 
\label{appdix:proof-thm-local-maximum-kmlec}

Suppose the inequality holds. Let $\hat{\Theta} \in M(\tau^*)$. Then we get
\begin{displaymath}
  \mathcal{L}(\tau^*,\Theta^*) = \mathcal{L}(\tau^*, \hat{\Theta})
  \geq \underset{\tau \in \mathcal{A}}{\max}\: \mathcal{L}(\tau, \hat{\Theta})
\end{displaymath}
So for any feasible direction $d$,
\begin{math}
  \left.
    d^T \frac{\partial \mathcal{L}(\tau,\hat{\Theta})}{\partial \tau}
  \right|_{\tau = \tau^*} \leq 0.
\end{math}
But this is true for all $\hat{\Theta} \in M(\tau^*)$ so
\begin{displaymath}
  \underset{\Theta \in M(\tau^*)}{\max}
  \left.
    d^T \frac{\partial \mathcal{L}(\tau, \hat{\Theta})}{\partial \tau}
  \right|_{\tau = \tau^*} \leq 0.
\end{displaymath}
But by Lemma~\ref{lem:DF-tau-star-d} the left hand side is $\mathrm{DF}(\tau^*;d)$,
so by Lemma~\ref{lem:tau-star-local-max-kmlec} $\tau^*$ is a local
maximum for k-MLE-c.

Now assume $\tau^*$ is a local maximum for k-MLE-c. For any feasible
direction $d$, by Lemma~\ref{lem:tau-star-local-max-kmlec}
$\mathrm{DF}(\tau^*; d) \leq 0$. Lemma~\ref{lem:DF-tau-star-d} ensures
\begin{math}
  \left.
    d^T \textstyle\frac{\partial \mathcal{L}(\tau,{\Theta})}{\partial \tau}
  \right|_{\tau = \tau^*} \leq 0
\end{math}
for any feasible $d$ and $\Theta \in M(\tau^*)$. Now consider a fixed
$\hat{\Theta} \in M(\tau^*)$. 
Recall also that
$\mathcal{L}(\tau, \hat{\Theta})$ is linear in $\tau$ by Property~P1.
These imply that
$\mathcal{L}(\tau^*, \hat{\Theta}) \geq \underset{\tau \in
  \mathcal{A}}{\max}\: \{ \mathcal{L}(\tau, \hat{\Theta}) \}$. Since
$\hat{\Theta}$ is arbitrary, hence (\ref{eq:F-geq-max-L}) holds and this
completes the proof.

\subsection{Proof of Theorem~\ref{thm:conv-singleton-local-maxima}}
\label{appdix:proof-thm-conv-singleton-local-maxima}

Since $(\tau^*, \Theta^*)$ is a partial maximum, then from the
definition
$\mathcal{L}(\tau^*, \Theta^*) \geq \underset{\tau \in
  \mathcal{A}}{\max}\: \mathcal{L}(\tau, \Theta^*)$. But when $M(\tau^*)$
is a singleton set this is exactly the local maximum condition of
Theorem~\ref{thm:local-maximum-kmlec}, thereby completing the proof.


\end{document}